\documentclass[10pt, a4paper]{article}

\usepackage[final]{lrec-coling2024} 
\newcommand{\ie}{\textit{i}.\textit{e}.}

\newcommand{\model}{Pro-QE}
\usepackage{booktabs}
\usepackage{multirow}
\usepackage{mathtools}

\title{Prompt-fused framework for Inductive Logical Query Answering}

\name{Zezhong Xu$^{\clubsuit\heartsuit}$, Peng Ye$^{\clubsuit}$, Lei Liang$^{\diamondsuit}$, Huajun Chen$^{\clubsuit\heartsuit\spadesuit}$, Wen Zhang$^{\clubsuit\heartsuit}$\sthanks{\ \ Corresponding author.}}
\address{$^\clubsuit$Zhejiang University, $^{\diamondsuit}$Ant Group\\
$^\spadesuit$Alibaba-Zhejiang University Joint Institute of Frontier Technology\\
$^\heartsuit$Zhejiang University-Ant Group Joint Laboratory of Knowledge Graph\\
        \{xuzezhong, yep, huajunsir, zhang.wen\}@zju.edu.cn\\
        leywar.liang@antgroup.com}

\abstract{
 Answering logical queries on knowledge graphs (KG) poses a significant challenge for machine reasoning.
 The primary obstacle in this task stems from the inherent incompleteness of KGs. Existing research has predominantly focused on addressing the issue of missing edges in KGs, thereby neglecting another aspect of incompleteness: the emergence of new entities. 
 Furthermore, most of the existing methods tend to reason over each logical operator separately, rather than comprehensively analyzing the query as a whole during the reasoning process.
 In this paper, we propose a query-aware prompt-fused framework named {\model}, which could incorporate existing query embedding methods and address the embedding of emerging entities through contextual information aggregation. Additionally, a query prompt, which is generated by encoding the symbolic query, is introduced to gather information relevant to the query from a holistic perspective.
 To evaluate the efficacy of our model in the inductive setting, we introduce two new challenging benchmarks. Experimental results demonstrate that our model successfully handles the issue of unseen entities in logical queries. Furthermore, the ablation study confirms the efficacy of the aggregator and prompt components.
  The code is available at \textit{https://github.com/yep96/Pro\_QE}.
 \\ \newline \Keywords{logical query, inductive reasoning, knowledge graph} }

\begin{document}

\maketitleabstract

\section{Introduction}
\begin{figure}[htbp]
    \centering
    \includegraphics[scale=0.65]{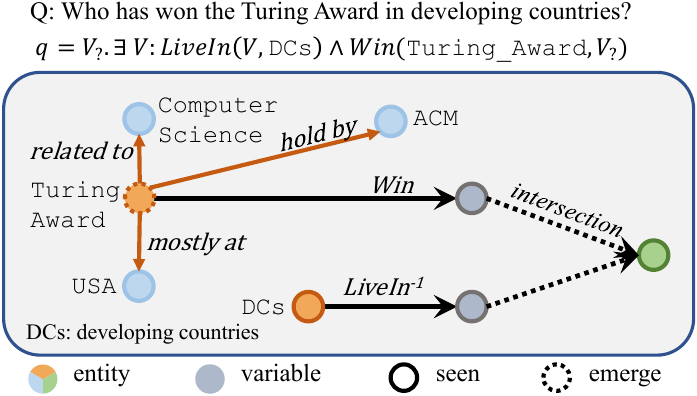}
    \caption{The example of an inductive logical query.}
    \label{fig:intro}
\end{figure}

Knowledge graphs (KGs) have garnered interest in recent years. Their applications span natural language processing, information retrieval, and data analytics.
The task of addressing logical queries on KGs encompasses multiple entities, relations, and logical operators. 
 For instance in Figure~\ref{fig:intro}, \textit{Who has won the Turing Award in developing countries?} could be translated into a logical query that requires traversing the KG to find the answer. In this query, the anchor nodes are the \textit{Turing Award} and \textit{Developing Countries}, while the target answer, such as \textit{Qizhi Yao}, represents the desired response.

Nonetheless, the existence of noise and incompleteness within KGs often leads to missing edges in the path from the anchor to the answer nodes, thereby impeding the retrieval of all pertinent answers through sub-graph matching. To tackle this challenge, a promising approach is to embed the query into a vector space, which facilitates the utilization of meaningful entity embeddings to handle missing or erroneous edges.
Typically, query embedding methods, such as GQE~\cite{GQE}, Q2B~\cite{Q2B}, BetaE~\cite{BetaE}, FuzzyQE~\cite{FuzzQE}, and GNN-QE~\cite{GNNQE1}, involve breaking down the query into simpler sub-queries, reasoning over sub-queries step by step. Subsequently, a distance function is employed to select entities that closely resemble the query results as the answer nodes. 

Despite the notable achievements of current query embedding methods in handling missing edges within KGs, there remain two unresolved issues.
The first issue pertains to the inductive query scenario. Existing query embedding models are primarily designed to handle scenarios where edges are missing in KG. However, KGs are dynamic and constantly evolving, which means new entities and relations emerging frequently. For instance, in Figure~\ref{fig:intro}, the anchor node \textit{the Turing Award} is a new entity, so its representation is randomly initialized.
We call this scenario the inductive setting, indicating queries may involve entities that were not present before. This task is challenging and necessitates advanced techniques and methodologies.
The inductive GNN-QE model~\cite{GNNQE2} is currently the only model that considers inductive scenarios. However, it still overlooks the coherence between the entire query and the inferred results, which leads us to the second unresolved issue.

The second issue concerns holistic query understanding in KGs. Complex KG queries involve multiple conditions and can exhibit intricate interdependencies. Most current models handle query reasoning by dividing queries into sub-queries. While these approach simplifies the problem and improves efficiency in some cases,  they hinder the model's ability to establish connections between different constraints from a holistic perspective.
For example, consider a query asking \textit{which US presidents were lawyers in American history}. A common approach involves querying for each logical operation separately: first querying for \textit{US presidents in history}, then for \textit{individuals who were lawyers}, and finally intersecting the results. However, this approach may retrieve non-Americans when querying for lawyers, unnecessarily expanding the scope. Conversely, when querying for lawyers, emphasizing nationality information can quickly retrieve relevant results without redundant operations.


This article aims to explore methodologies for conducting complex query reasoning in an inductive scenario. Our primary objectives include \textbf{Contextual Awareness}: The model should have the ability to obtain the representation of newly emerged entities by leveraging the representations of surrounding known entities.
\textbf{Query Awareness}: The model should comprehensively understand the query being posed and discern the role of each node within the query.
\textbf{Relevance Awareness}: The model should recognize that identical known entities can assume different roles in different contexts during the aggregation process. For example, the attributes of an esports athlete and a basketball athlete are distinct, even though both are athletes.
To address these objectives, we've developed a logical query-answering framework called \textit{\model} considering the three aspects.

We devised two benchmarks under the inductive scenario and conduct the experiments to evaluate the superior performance of our model.
  Furthermore, the ablation study supports the effectiveness of different module designs.
Our contributions can be summarized as follows:
\begin{itemize}
\item  We emphasize the significance of the inductive setting in the context of logical queries, highlighting the need for models to effectively handle new emerging entities.
\item  We advocate for a holistic approach to logical query processing, going beyond the decomposition of queries into sub-queries and considering the overall impact of the query on each step of the reasoning process.
\item  We establish new benchmarks for logical query processing under the inductive setting, conducting extensive experiments across various scenarios and achieving state-of-the-art results.
\end{itemize}

\section{Related work}
\subsection{Logical Query Answering}
Query answering in knowledge graphs has been widely studied in the literature.
 GQE(graph-query embedding)~\cite{GQE} first proposes to embed a query into a low-dimensional vector space and use the final embedding to find answers.
 Since GQE only supports intersection ($\wedge$) logical operator, Q2B (query to box)~\cite{Q2B} replaces the vector embedding with hyper-rectangles.
 To support the negation ($\neg$) operator, BetaE~\cite{BetaE} is proposed which models the query and entities using beta distributions.
 CQD (Continuous Query Decomposition)~\cite{CQD} applies beam search to an embedding model and could be trained with simple queries.
 ConE (Cone Embeddings)~\cite{ConE} proposes a new geometry model that embeds entities and queries using Cartesian products of two-dimensional cones.
 BiQE (Bidirectional Query Embedding)~\cite{BiQE} uses the transformer-based to incorporate the query context.
 TEMP (TypE-aware Message Passing)~\cite{TEMP} introduces a type-aware model making use of type information to assist logical reasoning.
 FuzzQE ~\cite{FuzzQE} satisfies the axiomatic system of fuzzy logic to reason.
 Q2P (Query2Particles)~\cite{Q2P} encodes each query into multiple vectors.
 GNN-QE~\cite{GNNQE1} concentrates on the interpretation of the variables along the query path.
Inductive GNN-QE~\cite{GNNQE2}(based on \cite{GNNQE1}) propose using NBFNet~\cite{NBFNet} as the one-hop(1p) projection to solve the inductive reasoning.
 ENeSy (Neural-Symbolic Entangled)~\cite{ENeSy} uses the neural and neural reasoning results to enhance each other to alleviate the problem of cascading error.


\subsection{Inductive Link Prediction}

Rule-based methods focus on mining rules to provide us with an interpretable reasoning process and can be applied to inductive link prediction. AMIE~\cite{AMIE} and AMIE+~\cite{AMIE+} use predefined metrics such as confidence and use three operations, including dangling atom, instantiated atom, and closing atom that add different types of atoms to incomplete rules.
Anyburl~\cite{AnyBURL} proposes a framework that can mine rules in an efficient way. 
Rudik~\cite{IEEE2018Rudik} can mine positive and negative rules, which can be used for link prediction tasks and other tasks.
More recently, differentiable rule learning methods attracted more attention, which can determine the rule confidence and structure of the rule. Neural-LP~\cite{yang2017differentiable} and DRUM~\cite{sadeghian2019drum} based on Tensorlog~\cite{tensorlog} generate the possibilities of different relationships in each step, and the symbolic rules can be parsed with the relations' weight.
Neural-Num-LP~\cite{2020_Num} extends Neural-LP to learn the numerical rules.
Ruleformer~\cite{ruleformer} utilizes context information to generate different rules according to a specific environment.
Neural Logic Inductive Learning~\cite{2019_NLIL} based-on transformer to mine non-chain-like rules which can extend the diversity rules.
More recent works exploit Graph Neural Networks, which can be applied to graph structures via the existing neighbors. \cite{OOKB} compute the embeddings of OOKB entities, exploiting the limited auxiliary knowledge provided at test time. 
LAN (Logic Attention Network)~\cite{LAN} aggregates neighbors with both rule-based attention and neural-based weights.
INDIGO~\cite{INDIGO} accomplishes inductive knowledge graph completion based on a graph convolutional network with pair-wise encoding.
Moreover, GraIL~\cite{GraIL}, CoMPILE~\cite{CoMPILE}, and TACT~\cite{TACT} learn the ability of relation prediction by subgraph extraction and GNNs independent of any specific entities. 
MorsE~\cite{MorsE} considers transferring universal, entity-independent meta-knowledge by meta-learning.

\section{Problem Formulation}

A KG can be represented as $\mathcal{G}=(\mathcal{V}, \mathcal{R}, \mathcal{T})$, where $\mathcal{V}$ denotes a set of entities, $\mathcal{R}$ signifies a set of relations, and $\mathcal{T}$ represents a set of triplets. Each triplet $r(e_i, e_j)$ in $\mathcal{T}$ is defined as a relation $r \in \mathcal{R}$ existing between entities $e_i$ and $e_j$ from the set $\mathcal{V}$. This notation illustrates that the relation $r$ holds between the entities $e_i$ and $e_j$. 

\subsection{Logical Query Answering}
As depicted in Figure~\ref{fig:intro}, a query comprises constants (utilized as anchor nodes), variables (employed as intermediate nodes and answer nodes), relation projections $r(a, b)$, signifying binary functions over variables or constants, and logic operators ($\vee$, $\wedge$, and $\neg$). Utilizing its disjunctive norm form~\cite{BetaE}, the query can be represented as a disjunction of several conjunctions, as shown below:
\begin{equation}
\label{FOL}
q[V_?] \coloneqq V_? . \exists V_1, V_2,\dots, V_k : c_1 \vee c_2 \vee \dots \vee c_n.
\end{equation}
where $c_i$ denotes either a single literal or a conjunction of multiple literals, i.e., $c_i = a_{i1} \wedge \dots \wedge a_{im}$, where $a_{ij}$ is an atom or the negation of an atom. The atoms can take the following forms: $r(e_a, V)$ or $\neg r(e_a, V)$ or $r(V^{\prime}, V)$ or $\neg r(V^{\prime}, V)$. Here, $e_a$ represents one of the constants (as there might be several anchor nodes), and $V, V^{\prime} \in {V_1, V_2, \dots, V_k, V_?}$ are the variables, with the constraint that an atom should satisfy $V \neq V^{\prime}$.
The operations above are precisely defined as follows:

\textbf{Relational Projection:} Given an entity set $\mathcal{S} \subseteq \mathcal{V}$ and a relation $r \in \mathcal{R}$, the relational projection operator returns an entity set $\mathcal{S}^{\prime}$ that includes entities connected to at least one of the entities in $\mathcal{S}$ via the relation $r$, i.e., $\mathcal{S}^{\prime} = {e^{\prime} \in \mathcal{V}| \exists r(e, e^{\prime}), e \in \mathcal{S}}$.

\textbf{Union:} Given sets of entities ${ S_1, S_2, \dots . S_n }$, the union operator returns a new set $\mathcal{S}^{\prime}$ that contains the entities appearing in any one of the sets, i.e., $\mathcal{S}^{\prime} = \bigcup_{i=1}^{n} S_i$.

\textbf{Intersection:} Given sets of entities ${ S_1, S_2, \dots, S_n }$, the intersection operator returns a new set $\mathcal{S}^{\prime}$ that contains the entities belonging to the intersection of these sets, i.e., $\mathcal{S}^{\prime} = \bigcap_{i=1}^{n} S_i$.

\textbf{Complement:} Given a set of entities $\mathcal{S}$, this operator returns a new entity set $\mathcal{S}^{\prime}$, which is the complement of $\mathcal{S}$, i.e., $\mathcal{S}^{\prime} = \mathcal{V} - \mathcal{S}$.
 

\subsection{Inductive Setting}
The inductive setting, as explored in this paper, addresses scenarios in which there might be new entities that only appear in the validation and test sets. We encounter two distinct entity sets: $\mathcal{V}_{train}$ and $\mathcal{V}_{test}$. The former represents the entity set in the training set, while the latter represents all the entities in the test process. The inclusion of new entities in the test query introduces a unique set of challenges, notably the necessity to obtain representations of these new entities.
\begin{figure*}[htbp]
\centering
  \includegraphics[scale=0.54]{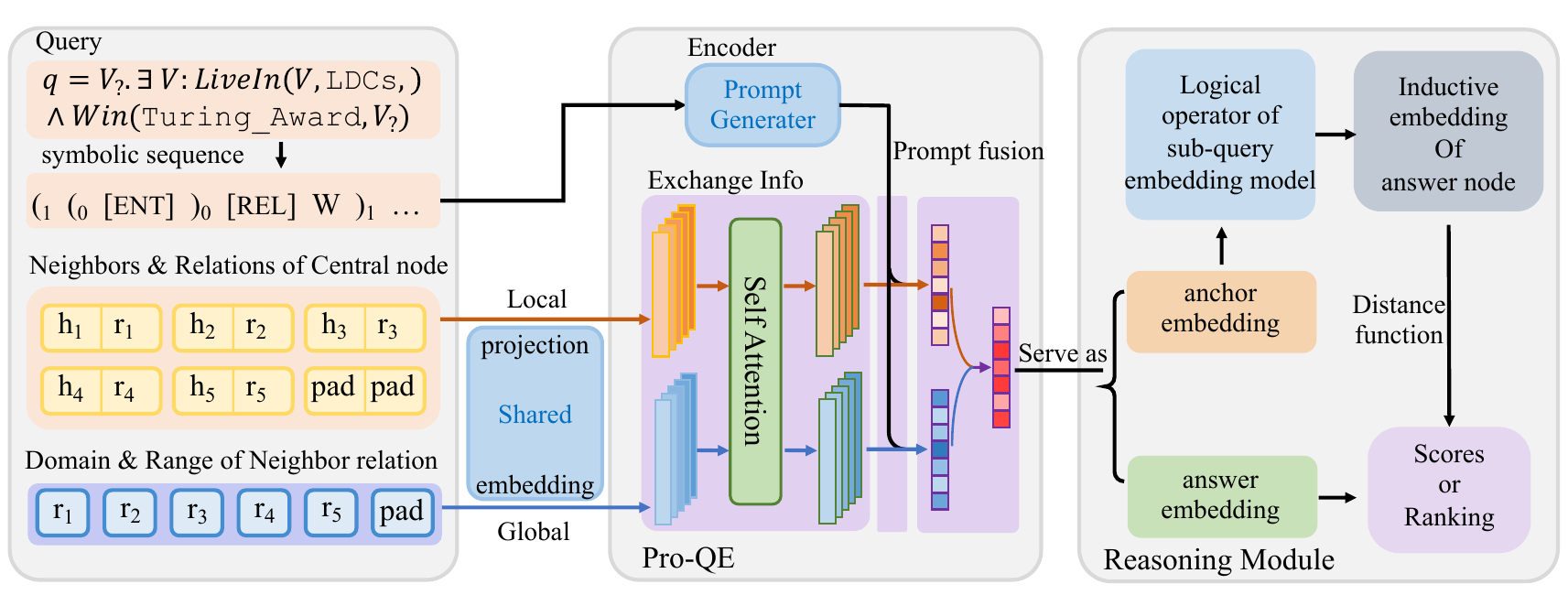}
  \caption{Framework of the proposed {\model} model.}
  \label{fig:framework}
\end{figure*}
Additionally, new triplets associated with emerging entities could result in an overlap between the training and test graphs. This overlap may lead to an increase in the number of observed answers to the same query in the training set, making it challenging to distinguish between novel and existing answers.
We have classified the queries based on the location of new entities in the query:
1. New entities only appear in the anchor nodes.
2. New entities only appear in the answer set.
3. New entities appear in both the anchor nodes and the answer set.
It is worth noting that there exists another type of query where new entities only appear in the reasoning process. Since the answer set can still be obtained using only the training data, we do not consider this type of query.

\section{Methodology}


In this section, we introduce the comprehensive architecture of {\model}, which is depicted in Figure~\ref{fig:framework}.
Specifically, we leverage contextual information to obtain an appropriate representation of entities. During this process, we utilize neighbor nodes and relations to capture local information, and relation domains and ranges to capture global information. To enhance relevance awareness of different neighbors, we incorporate the attention mechanism.
Additionally, we integrate the entire query information from a holistic perspective, which enables the model to better comprehend the overall query and its impact on each step of reasoning.

\subsection{Inductive Representation Learning}
To obtain a suitable representation of an entity, even in the case of it being an emerging entity, we employ a context-informed approach. During this process, we conscientiously consider both local and global information to derive a comprehensive representation. To ensure heightened relevance awareness, we leverage a self-attention mechanism for seamless information exchange. 

 \textbf{Local Information.}
 An entity's representation can be determined by its contextual information, \ie, the subgraph in KG. We define an input embedding denoted as $\mathbf{e}^{I}$, which serves as input to derive representations of other entities connected to it. We could get the output embedding of the central entity using the input embedding of its neighbors, and the output embedding could be utilized for reasoning. In this paper, we consider the subgraph composed of the one-hop neighboring nodes.
 
 Consider a central entity $e_{i}$ with its one-hop neighboring entity set denoted as $\mathcal{N}_{e}(i)$ and relation set as $\mathcal{N}_{r}(i)$. 
 For each entity $e_{j} \in \mathcal{N}_{e}(i)$ and relation $r_j \in \mathcal{N}_{r}(i)$ linking $e_i$ and $e_j$, the input embedding $\mathbf{e}_{j}^{I}$ and relation embedding $\mathbf{r}_{j}$ are randomly initialized.
 We apply a projection operator to $e_{j}$ with relation $r$, to derive the projected representation $\mathbf{e}_{j}^{P}$. The projection could be any method that could make the link prediction. For instance, if we employ GQE~\cite{GQE} or FuzzQE~\cite{FuzzQE}, the projection operator is applied in the following manner:
\begin{equation}
    \label{Projection_GQE}
    \mathbf{e}_{j}^{P} = P(\mathbf{e}_{j}^{I}) = \mathbf{e}_{j}^{I} + \mathbf{r_{j}}
\end{equation}
\begin{equation}
    \label{Projection_fuzzQE}
    \mathbf{e}_{j}^{P} = P(\mathbf{e}_{j}^{I}) = \mathbf{g}( \textbf{LN}(\mathbf{W}_{r_{j}}\mathbf{e}_{j}^{I} + \mathbf{b}_{r_{j}}))
\end{equation}
 where the $\textbf{g}$ is a mapping function, and \textbf{LN} is Layer Normalization function. $\mathbf{r_{j}}$ and $\mathbf{W}_{r_{j}}$ are both relation parameters according to the projection function, which could be seen as $\mathbf{r}_{j}$.
 After performing the projection on each entity-relation pair, we derive a set of representations, in which each vector partially reflects the information of the central entity. 
 
\textbf{Global Information.}
 Since all of this information is exclusively derived from the subgraph, it may contain biases and noise. If the central entity is linked to a few-shot entity (not an emerging entity), the local representation derived from it may lack accuracy. Thus, it is necessary to explore more robust representations on a larger scale. In this paper, we consider the domain and range of relations, as entities which are related to the same domain or range are likely to exhibit similar features.
 
\begin{figure*}[htbp]
    \centering
    \includegraphics[scale=0.38]{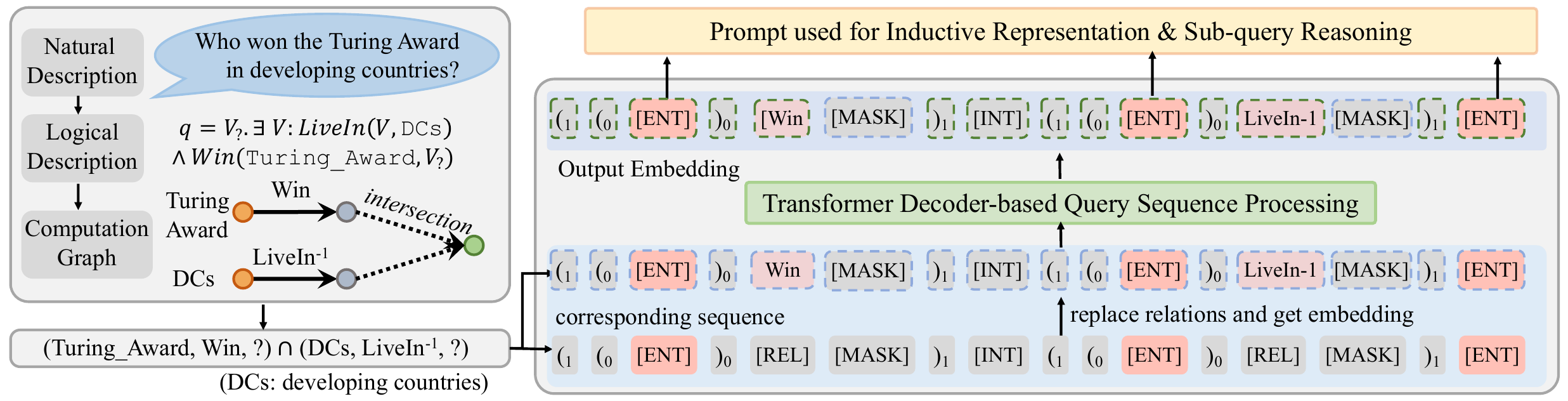}
    \caption{Architecture of the query prompt processing system.}
    \label{fig:prompt}
\end{figure*}
For each relation $r$, we define two entity sets based on the entities related to it: one set encompasses the entities serving as the head of the relation, signifying the domain of the relation. The other set contains the tail entity of the relation, representing the range of the relation. We initialize a representation $\textbf{e}_{do(r)}$ and $\textbf{e}_{ra(r)}$ for each such set to encapsulate all the entities within it. For a central entity and its related relation set, the corresponding embedding set $\textbf{e}_{do(r_{1})}...\textbf{e}_{do(r_{m})}, \textbf{e}_{ra(r_{1})}... \textbf{e}_{ra(r_{n})}$ could be perceived as reflecting information of $e_{i}$, where $m+n$ is the number of relations connected with $e_{i}$.

\textbf{Information Exchange and Aggregation.}
Now we have obtained an embedding set, $\mathbf{E}^{L}$ and $\mathbf{E}^{G}$, from both local and global perspectives, but have not established any associations between these embeddings. In reality, there exist direct or implicit connections between different neighbors. For example, two sports players may share a link to the same entity \textit{sports player}, but their sport is \textit{esports} and \textit{basketball}, respectively. Consequently, even though both are associated with the \textit{sports player} entity, their embeddings may vary when employed in subsequent tasks.

 To facilitate information exchange between neighboring nodes, we utilize the attention module:
\begin{equation}
    \label{attention_local}
    \mathbf{E}^{L \prime} = \texttt{Attn}(\mathbf{W}_{q}^{L}\mathbf{E}^{L}, \mathbf{W}_{k}^{L}\mathbf{E}^{L}, \mathbf{W}_{v}^{L}\mathbf{E}^{L})
\end{equation}
\begin{equation}
    \label{attention_global}
    \mathbf{E}^{G \prime} = \texttt{Attn}(\mathbf{W}_{q}^{G}\mathbf{E}^{G}, \mathbf{W}_{k}^{G}\mathbf{E}^{G}, \mathbf{W}_{v}^{G}\mathbf{E}^{G})
\end{equation}
 where the $\mathbf{W}_{q}^{L}$, $\mathbf{W}_{k}^{L}$ and $\mathbf{W}_{v}^{L}$ are the metrics used for modeling the Query, Key, and Value for the self-attention mechanism for Local embedding. $\mathbf{W}_{q}^{G}$, $\mathbf{W}_{k}^{G}$ and $\mathbf{W}_{v}^{G}$ are the same for Global information.

 The \texttt{Attn} represents the self-attention with the Query, Key, and Value, which are denoted as Q, K, and V as input:
 \begin{equation}
    \label{self-attention}
    \texttt{Attn}(Q,K,V) = \texttt{Softmax}(\frac{QK^{T}}{\sqrt{d}})V
 \end{equation}
 Here, Q, K, and V represent the Query, Key, and Value inputs for this attention layer.

 However, we have used the context-aware and relevance-aware information, but lacking query-specific awareness. To address this limitation, we introduce a prompt mechanism.

\subsection{Logical Query Prompt}

 As previously mentioned, in the sub-query model, each reasoning step lacks a comprehensive understanding of the overall query. We employ a method that transforms the entire query into a symbolic query statement and utilizes the decoder architecture of the transformer\cite{transformer} to handle the query sequence.

 The figure depicted in Figure~\ref{fig:prompt} illustrates the comprehensive process of this step. We utilize the symbol [ENT] to represent anchor nodes and answer nodes, while [MASK] is employed to signify intermediate entities during the reasoning process. [r\_id] is used to denote specific relations utilized in the projection operation, while [INT], [UNI], and [NEG] stand for intersection, union, and negation operators, respectively. Additionally, to preserve the order of different operations within a symbol sequence, we adopt [BCK\_L] and [BCK\_R] to represent brackets.

 Next, to get the sequence, we begin with each [ENT] symbol representing an anchor node and enclose it within a pair of brackets [BCK\_L${0}$] and [BCK\_R${0}$]. If the subsequent operation involves projection or negation, we append the [r\_id] or [NEG] operator, followed by a [MASK] symbol signifying an intermediate node in the computational graph. This procedure is repeated until an intersection or union operation is encountered, and then the elements requiring intersection or union are enclosed within higher-level brackets, i.e., [BCK\_L${1}$] and [BCK\_R${1}$] and so on.
 Finally, we add [MASK] at the end of the sequence to represent the operation itself and the result after the operation. If a union or intersection operation is required later, a pair of brackets representing the branch should be added outside the sequence representing that branch. Finally, an [ENT] token will be added at the end of the sequence to represent the answer node.

 Capitalizing on the transformer model's exceptional ability to process sequence data, we exploit the transformer's decoder architecture to handle the symbol sequences generated through the aforementioned method. Specifically, we convert the symbol sequence into corresponding embeddings. It is crucial to note that each embedding of the [r\_id] specific to a particular relation is distinct.
 \begin{equation}
    \label{prompt_generation}
    \mathbf{O} = \texttt{Decoder}(\texttt{Seq}(q))
 \end{equation}
 Here, \texttt{Seq}(q) signifies the symbol sequence generated based on query $q$, which is utilized as input to the decoder, and the resulting output sequence is embedded as $\mathbf{O}$.

\begin{figure*}[htb]
\centering
  \includegraphics[scale=0.57]{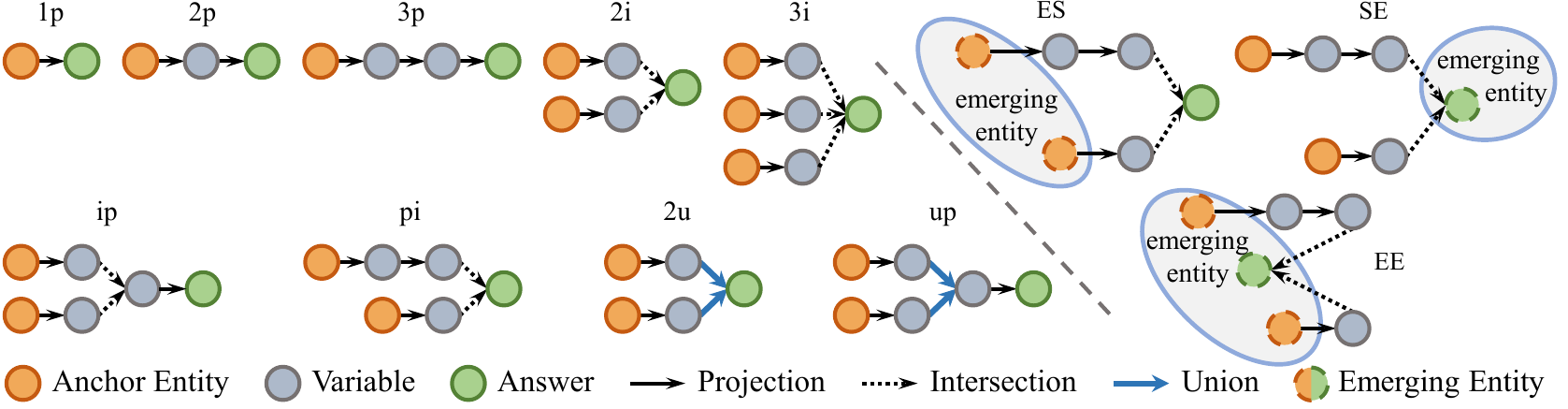}
  \caption{The query structures and query types used for grounding the specific queries on the KGs. The p, i, and u denote the projection, intersection, and union operator, respectively.}
  \label{fig:query_structure}
\end{figure*}

For each vector in $\mathbf{E}^{L \prime}$ and $\mathbf{E}^{G \prime}$, we compute the dot product with the prompt to establish the correlation between the embedding and the prompt. Subsequently, this correlation is employed as the weight coefficient during the aggregation of the anchor node's representation. For the local information:
  \begin{equation}
    \label{weight}
    \alpha_{ij} = \textbf{p}_{i} \cdot \textbf{e}_{j}^{\prime}, \textbf{e}_{j}^{\prime} \in \mathbf{E}^{L \prime}
 \end{equation}
\begin{equation}
    \label{final_embedding}
    \textbf{e}_{i}^{L} = \sum_{j=0}^{\mathcal{N}_{e}(i)} \alpha_{ij}  \textbf{e}_{j}^{\prime}
 \end{equation}
 For the global embedding, we use the same way to get $\textbf{e}_{i}^{G}$, and the final embedding $\textbf{e}_{i} = (\textbf{e}_{i}^{G}+\textbf{e}_{i}^{L})/2$.

\subsection{Reasoning and Training}


To train the model, let's consider a query $q$ and the set of candidate answers $\mathcal{V}_{q}$, we choose a number of negative samples, denoted as $e^{\prime}$. The embedding of the anchor nodes and the samples could be obtained using the methods described before. The query embedding result could be obtained using existing models like GQE~\cite{GQE} with the anchor node(s) embedding(s).
Our objective is to optimize the embeddings in such a way that the positive sample $\mathbf{e}$ is closer to the query embedding $\mathbf{q}$, while the negative samples $\mathbf{e}_{i}^{\prime}$, where $i=1\dots k$, are pushed farther away from $\mathbf{q}$.
To achieve this goal, we calculate the loss using the distance between the query embedding and each of the positive and negative samples:
\begin{align}
    \label{loss}
    L = -\text{log} \sigma(\gamma - \texttt{D}(\mathbf{q}, \mathbf{e})) -
    \frac{1}{k}\sum_{i=1}^{k}\text{log} \sigma(\texttt{D}(\mathbf{q}, \mathbf{e^{\prime}_{i}}) - \gamma)
\end{align}
where $\gamma$ is a scalar margin. \texttt{D} is the distance function. 
The batch loss for training samples can be formulated as follows:
\begin{align}
    \label{loss_batch}
    L_{b} = \sum_{i=1}^{l} \frac{1}{w} L_{i}
\end{align}
where $l$ denotes the total number of samples in a given batch, $L_{i}$ represents the loss calculated according to Equation~\ref{loss} for query $q_{i}$, and $\frac{1}{w}$ denotes the weight coefficient for the sample's loss. In this article, we adopt $\frac{1}{w}=\sqrt{|\mathcal{V}_{q_{i}}|}$ as the weight coefficient.

\section{Experiment}

In this section, we provide comprehensive details of our experiments. Firstly, we curated two inductive logical query benchmarks.
Secondly, we established two baseline models to serve as a point of comparison in our experiments.
Lastly, we performed ablation experiments to analyze our model, to verify the effectiveness of each component.
\begin{table}
\centering
\caption{Statistics of the constructed datasets. Note that it contains all 9 query structures.}
\label{tab:dataset}
\resizebox{\textwidth/2}{!}{ 
\begin{tabular}{ccccc} 
\toprule
Dataset         & Train   & EE(Valid / Test) & ES(V/T) & SE(V/T)  \\ 
\hline
FB15k-237 & 900,000 & 22,500  & 22,500  & 22,500   \\
NELL-995  & 585,000  & 18,000  & 18,000  & 18,000   \\
\bottomrule
\end{tabular}
}
\end{table}

\subsection{Benchmarks Construction}

We establish the required benchmarks using the FB15k-237~\cite{FB15k-237} and NELL-995~\cite{NELL995} datasets, which are the same as the datasets used in the previous works of logical query answering, and Table~\ref{tab:dataset} shows the statistics of the datasets:

\begin{table*}[htb]
\centering
\caption{The inductive logical query answering results of different structures and types on FB15k-237.}
\label{tab:results1}
\resizebox{\textwidth}{!}{ 
\begin{tabular}{cccccccccccccccc} 
\toprule
\multirow{2}{*}{\begin{tabular}[c]{@{}c@{}}\textbf{Query}\\\textbf{Structure}\end{tabular}} & \multicolumn{3}{c}{\textbf{Mean(GQE)}}  &  & \multicolumn{3}{c}{\textbf{Feature(GQE)}} &  & \multicolumn{3}{c}{\textbf{\model(GQE)}} &  & \multicolumn{3}{c}{\textbf{\model(Q2B)}}  \\ 
\cline{2-16}
                                                                                            & \textbf{EE} & \textbf{ES} & \textbf{SE} &  & \textbf{EE} & \textbf{ES} & \textbf{SE}   &  & \textbf{EE} & \textbf{ES} & \textbf{SE}                     &  & \textbf{EE}   & \textbf{ES}   & \textbf{SE}                  \\ 
\hline
\textbf{1p}                                                                                 & 20.0        & 18.2        & 12.3         &  & 12.2        & 12.7        & 14.4          &  & 26.1        & 25.2        & 34.0                             &  & 28.1          & 28.4          & 37.8                         \\
\textbf{2p}                                                                                 & 12.9         & 15.4        & 11.2         &  & 7.6         & 11.2        & 11.4          &  & 17.2        & 24.8        & 25.0                             &  & 19.0          & 27.4          & 24.6                          \\
\textbf{3p}                                                                                 & 9.5         & 15.6         & 9.7         &  & 5.9         & 12.0        & 9.5           &  & 14.7        & 21.4        & 20.5                             &  & 15.5          & 25.2          & 19.5                          \\
\textbf{2i}                                                                                 & 12.9        & 20.4        & 12.0         &  & 12.9        & 31.1        & 12.7          &  & 19.9        & 34.4        & 31.5                             &  & 21.0          & 34.2          & 35.8                          \\
\textbf{3i}                                                                                 & 14.1        & 23.6        & 11.2         &  & 13.7        & 40.7        & 12.5          &  & 24.6        & 38.7        & 32.4                             &  & 28.2          & 40.3          & 40.2                          \\
\textbf{pi}                                                                                 & 10.3         & 17.5        & 8.8         &  & 10.1        & 26.2        & 11.2          &  & 15.6        & 30.5        & 25.8                             &  & 17.0          & 32.2          & 28.6                          \\
\textbf{ip}                                                                                 & 13.8        & 25.0        & 11.9         &  & 11.6        & 32.7        & 11.2          &  & 17.4        & 34.1        & 21.9                             &  & 19.8          & 38.0          & 22.4                          \\
\textbf{2u}                                                                                 & 9.4         & 29.5        & 12.6         &  & 7.6         & 35.4        & 11.5          &  & 11.3        & 43.8        & 23.6                             &  & 12.1          & 45.5          & 25.8                          \\
\textbf{up}                                                                                 & 12.6        & 31.0        & 12.8         &  & 9.1         & 34.5        & 12.5          &  & 16.1        & 42.6        & 22.9                             &  & 16.8          & 35.1          & 21.5                          \\ 
\hline
\textbf{avg}                                                                                & 12.8        & 21.8        & 11.4         &  & 10.1        & 26.3        & 11.9          &  & 18.1        & 32.8        & 26.4                             &  & \textbf{19.7} & \textbf{35.1} & \textbf{28.5}                          \\
\bottomrule
\end{tabular}
}
\end{table*}

\begin{table*}[htb]
\centering
\caption{The inductive logical query answering results of different structures and types on NELL-995.}
\label{tab:results2}
\resizebox{\textwidth}{!}{ 
\begin{tabular}{cccccccccccccccc} 
\toprule
\multirow{2}{*}{\begin{tabular}[c]{@{}c@{}}\textbf{Query}\\\textbf{Structure}\end{tabular}} & \multicolumn{3}{c}{\textbf{Mean(GQE)}}  &  & \multicolumn{3}{c}{\textbf{Feature(GQE)}} &  & \multicolumn{3}{c}{\textbf{{\model}(GQE)}}        &  & \multicolumn{3}{c}{\textbf{{\model}(Q2B)}}               \\ 
\cline{2-16}
                                                                                            & \textbf{EE} & \textbf{ES} & \textbf{SE} &  & \textbf{EE} & \textbf{ES} & \textbf{SE}   &  & \textbf{EE} & \textbf{ES} & \textbf{SE} &  & \textbf{EE}   & \textbf{ES}   & \textbf{SE}    \\ 
\hline
\textbf{1p}                                                                                 & 4.0         & 6.7         & 15.6        &  & 4.7         & 8.5         & 25.2          &  & 7.2         & 13.5        & 35.8        &  & 8.6           & 14.8          & 40.2           \\
\textbf{2p}                                                                                 & 18.24        & 17.0        & 10.5         &  & 10.8        & 12.2        & 12.7          &  & 29.8        & 36.2        & 25.7        &  & 31.8          & 38.2          & 27.3           \\
\textbf{3p}                                                                                 & 4.4         & 10.2         & 6.9         &  & 5.2         & 8.1         & 9.7           &  & 11.0         & 17.0        & 16.8        &  & 12.0           & 19.8          & 18.8           \\
\textbf{2i}                                                                                 & 9.1         & 23.7        & 10.8         &  & 10.1        & 30.2        & 14.8          &  & 15.0         & 38.4        & 24.3        &  & 13.7          & 41.3          & 31.1           \\
\textbf{3i}                                                                                 & 8.7         & 35.7        & 6.2         &  & 9.1         & 37.1        & 10.6          &  & 14.4         & 45.5        & 18.3         &  & 13.6           & 50.6          & 25.2           \\
\textbf{pi}                                                                                 & 9.3         & 20.1        & 11.9         &  & 9.1         & 26.1        & 18.3          &  & 18.8        & 34.1        & 32.6        &  & 19.3          & 37.9          & 40.9           \\
\textbf{ip}                                                                                 & 24.3        & 30.3        & 17.5        &  & 17.8        & 42.1        & 14.7          &  & 38.7        & 57.3        & 32.9        &  & 37.0          & 56.6          & 32.5           \\
\textbf{2u}                                                                                 & 8.9         & 29.6        & 11.0         &  & 8.5         & 38.6        & 14.2          &  & 11.7        & 46.5        & 23.0        &  & 17.2          & 53.9          & 30.2           \\
\textbf{up}                                                                                 & 22.9        & 32.8        & 15.5        &  & 15.0        & 50.9        & 13.2          &  & 33.5        & 57.3        & 27.8        &  & 33.4          & 58.7          & 26.5           \\ 
\hline
\textbf{avg}                                                                                & 12.2        & 22.9        & 11.8         &  & 10.0        & 28.2        & 14.8          &  & 20.0        & 38.4        & 26.4        &  & \textbf{20.7} & \textbf{41.3} & \textbf{30.3}  \\
\bottomrule
\end{tabular}
}
\end{table*}

\textit{Random Sampling of Unseen Entities and Triplet Splitting}: We employ random selection to extract 20\% of entities from the entity set $\mathcal{V}$, forming the set of emerging entities, denoted as $\mathcal{V}_{test}$. The remaining entities are considered seen entities and are represented as $\mathcal{V}_{train}$.
For a triplet $r(h, t) \in \mathcal{T}$, if any entity is present in the emerging entity set, i.e., $h \in \mathcal{V}_{test}$ or $t \in \mathcal{V}_{test}$, it is an auxiliary triplet. Otherwise, it is considered a training triplet. The triplet set $\mathcal{T}$ can be split into an auxiliary triplet set $\mathcal{T}_{a}$ and a training triplet set $\mathcal{T}_{t}$.

\textit{Ground the queries.}
The training triplet set $\mathcal{T}_{t}$ is utilized to construct the training graph $\mathcal{G}_{t}$. The entire graph $\mathcal{G}$ is used during the validation and testing steps.
For different query structures in Figure~\ref{fig:query_structure}, we ground the query in $\mathcal{G}_{t}$ to create the training query, where both the anchor and answer entities are exclusively seen entities. In contrast, to generate the validation and test queries, we ground the query in the graph $\mathcal{G}$, resulting in the anchor and answer entities potentially including unseen entities.

\textit{Classify the queries.} The validation and test queries may contain emerging entities, so they could be categorized into four types: (1) Emerging anchor entities and emerging answer entities, (2) Emerging anchor entities and seen answer entities, (3) Seen anchor entities and emerging answer entities, and (4) Seen anchor entities and seen answer entities, denoted as \textbf{EE, ES, SE, SS} type queries. We use the first three types of queries to evaluate the inductive query reasoning ability since the last type can be answered only using the training graph.

\subsection{Experimental Configurations}
\subsubsection{Evaluation Protocol}
 The evaluation of EE and ES type queries involves considering all entities in $\mathcal{V}_{q}$ obtained by traversing the graph because the anchor entity in these queries is an unseen entity. In contrast, for SE type queries, the answer set may contain both seen and unseen entities, and only the entities belonging to $\mathcal{V}_{test}$ are evaluated. The answer set $A = \mathcal{V}_{q}$ and $ \mathcal{V}_{q} \backslash \mathcal{V}_{test}$ for $q \in EE || ES$ and  $q \in SE$, respecitively.
 We calculate the Mean Reciprocal Rank (MRR):
 \begin{equation}
 \label{metrics1}
    MRR(q) = \frac{1}{|A|} \sum_{v \in A} \frac{1}{R(v)}, .
\end{equation}
 
where $R(v)$ means the rank of $v$ in all candidate entities according to our results.

\subsubsection{Baselines}
In our study on inductive logical query reasoning, we incorporated our proposed model, {\model}, with existing query embedding methods on KG such as GQE~\cite{GQE} and Q2B~\cite{Q2B}.
To provide a baseline for comparison, we implemented two alternative methods. The first method involves computing the average representation of neighbor embeddings as the central embedding.
The second method relies on evaluating entity features based on their neighbor relations and entities. Specifically, we train the embeddings on the training graph for the seen entities. During the test process, we compare the emerging entity with all the seen entities and use the embedding of the seen entity that shares the most common feature (neighbor relation and entity) with the emerging entity as the anchor node.

\subsubsection{Experimental Setup}
Our model was implemented using the PyTorch and trained on RTX3090. We employed the ADAM optimizer. The learning rate was set to be 0.0001, and it was halved at the halfway point of the training steps to aid the model's convergence.
We set the embedding dimension to 512 for entities, relation domains, relation ranges, and relations. The training batch sizes were set to 32, and the negative sample size was also set to 32.
For the loss function, the margin parameter ($\gamma$) was set to be 24.
We set the number of neighbors to 32 for FB15k-237 and 5 for NELL-995 datasets during inference, since it could overlap all the neighbors of 85\% entities in the datasets.


\subsection{Main Results}
 We conducted a comprehensive comparison of our model's performance against two baseline models: Mean and Feature matching, both of which are based on GQE~\cite{GQE}.
 Additionally, we enhanced our model's capabilities by incorporating query embedding techniques from both GQE~\cite{GQE} and Q2B~\cite{Q2B}. It is important to note that the query embedding techniques employed in this investigation were designed to handle only positive queries. Integrating models equipped with negation operations would extend the overall model's capabilities to manage the negation operator.

The results shown in Table~\ref{tab:results1} and Table~\ref{tab:results2}, revealed that our model consistently outperformed the baselines.
Notably, our model achieved an average improvement of approximately 6.9\%, 13.3\%, and 17.1\% (equivalent to 53.9\%, 61.0\%, and 150\%, respectively) on EE, ES, and SE query types, compared to the Mean model, which also utilizes context information, for the FB15k-237 dataset. For the NELL-995 dataset, our model achieved an MRR improvement of 8.5\%, 18.4\%, and 18.5\% (equivalent to 70.0\%, 80.3\%, and 156.8\% respectively) for the three different query types, respectively.

Furthermore, we made the observation that our model based on Q2B demonstrated the most superior overall performance compared to our model based on GQE. Specifically, our results indicated that the performance of our model based on Q2B outperformed that of our model based on GQE, with an improvement of 1.6\%, 2.3\%, and 2.1\% on FB15k-237, and 0.7\%, 2.9\%, and 3.9\% on NELL-995, respectively.
This finding is consistent with the relative performance of the two models reported in the original source paper~\cite{Q2B}.
Additionally, the results highlighted that queries of type EE were the most challenging.


\begin{table}
\centering

\caption{Ablation study results on NELL-995.}
\resizebox{\textwidth / 2}{!}{ 
\begin{tabular}{cccccccccc} 
\toprule
\multirow{2}{*}{} & \multicolumn{3}{c}{\begin{tabular}[c]{@{}c@{}}\textbf{w/o}\\\textbf{Global}\end{tabular}} & \multicolumn{3}{c}{\begin{tabular}[c]{@{}c@{}}\textbf{w/o }\\\textbf{Info Exchange}\end{tabular}} & \multicolumn{3}{c}{\begin{tabular}[c]{@{}c@{}}\textbf{w/o }\\\textbf{Query Prompt}\end{tabular}}  \\ 
\cline{2-10}
                  & \textbf{EE} & \textbf{ES} & \textbf{SE}                                                   & \textbf{EE} & \textbf{ES} & \textbf{SE}                                                           & \textbf{EE} & \textbf{ES} & \textbf{SE}                                                           \\ 
\hline
\textbf{1p}       & 6.9         & 12.7        & 34.3                                                          & 6.4         & 10.8        & 31.9                                                                  & 5.0         & 8.2        & 25.0                                                                  \\
\textbf{2p}       & 30.5        & 38.0        & 25.1                                                          & 32.9        & 41.4        & 27.3                                                                  & 24.4        & 23.9        & 18.2                                                                  \\
\textbf{3p}       & 7.5         & 15.2        & 15.0                                                          & 9.9         & 13.9        & 15.9                                                                  & 6.9         & 10.9        & 12.1                                                                  \\
\textbf{2i}       & 12.8         & 34.2        & 22.8                                                          & 11.6         & 29.6        & 21.6                                                                  & 11.0         & 26.9        & 16.9                                                                  \\
\textbf{3i}       & 12.5         & 37.2        & 17.5                                                           & 11.0         & 36.5        & 15.0                                                                   & 9.9         & 34.2        & 10.9                                                                   \\
\textbf{pi}       & 17.4        & 31.4        & 32.5                                                          & 17.5        & 28.7        & 32.0                                                                  & 14.7        & 24.5        & 22.4                                                                  \\
\textbf{ip}       & 38.9        & 57.5        & 32.6                                                          & 34.8        & 54.2        & 30.6                                                                  & 32.6        & 45.7        & 25.7                                                                  \\
\textbf{2u}       & 10.8        & 44.4        & 20.4                                                          & 9.2         & 37.2        & 19.1                                                                  & 9.3        & 36.0        & 15.3                                                                  \\
\textbf{up}       & 34.2        & 57.0        & 26.4                                                          & 31.7        & 54.2        & 26.6                                                                  & 28.6        & 45.4        & 21.2                                                                  \\
\bottomrule
\end{tabular}
}
\label{tab:ablation}
\end{table}

\subsection{Ablation Study}

To evaluate the effectiveness of the different modules, we conducted ablation experiments on the NELL-995 dataset.

\begin{figure}[h]
    \centering
    \includegraphics[scale=0.35]{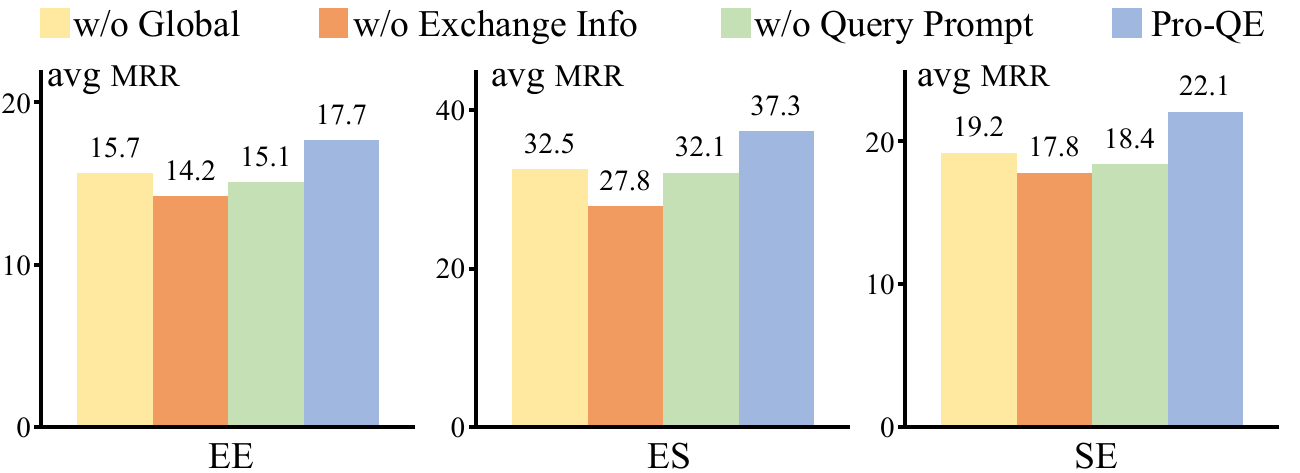}
    \caption{The average MRR results on NELL-995 of {\model} without different module.}
    \label{fig:ablation}
\end{figure}

To ensure that our results were consistent, we used the same logical query reasoning module, GQE~\cite{GQE}. Our results are summarized in Table~\ref{tab:ablation}. We also compared the average MRR results of different models in Figure~\ref{fig:ablation}.

The results demonstrated that when disregarding \textit{global information}, the performance of inductive reasoning was significantly hindered, showing a decline of approximately 0.9\%, 2.0\%, and 1.2\% in absolute values (equivalent to 4.7\%, 5.5\%, and 4.8\% relatively) for the average MRR metrics of the three query types, respectively, suggesting that local information, which can potentially be perturbed and biased, tends to limit the model's potential.

Moreover, the results revealed a noteworthy improvement in the model's performance when we incorporated the information exchange mechanism. There was an average increase of approximately 1.6\%, 4.3\%, and 2.0\% (or 8.7\%, 12.6\%, and 8.2\%, respectively) for the three query types (EE, ES, and SE). This indicates that the interaction between neighbors can enhance the quality of aggregation.
Finally, the most significant improvement is achieved with the \textit{query prompt} module.
The MRR results improved by about 4.2\%, 10.0\%, and 7.7\% (equivalent to relative improvements of 26.6\%, 35.2\%, and 41.2\%) for the EE, ES, and SE query types, respectively. This improvement was particularly pronounced for the \textit{3p} and \textit{pi} queries, which has the longest length from the anchor node to the answer node, and this means that there is more overall information hidden in the query.

\section{Conclusion}
In this paper, we address the problem of inductive logical query answering with emerging entities and present a novel framework, called {\model}. The model adheres to three key principles - utilizing both local and global contexts, incorporating a self-attention mechanism for information exchange, and employing a query prompt encoder to obtain an accurate representation of unseen entities.
The results of the experiments conducted on these benchmarks demonstrate the efficacy of the proposed model and principles. 
Future work may include introducing new relations in inductive settings.

\section*{Acknowledgements}
This work is founded by National Natural Science Foundation of China (NSFC623062
76/NSFCU23\\B2055/NSFCU19B2027/NSFC91846204), Zhejiang Provincial Natural Science Foundation of China (No. LQ23F020017), Ningbo Natural Science Foundation (2023J291), Yongjiang Talent Introduction Programme (2022A-238-G),  Fundamental Research Funds for the Central Universities (226-2023-00138).

\bibliographystyle{lrec-coling2024-natbib}
\bibliography{lrec-coling2024-example}

\begin{thebibliography}{0}
\expandafter\ifx\csname natexlab\endcsname\relax\def\natexlab#1{#1}\fi

\end{thebibliography}


\begin{thebibliography}{33}
\expandafter\ifx\csname natexlab\endcsname\relax\def\natexlab#1{#1}\fi

\bibitem[{Arakelyan et~al.(2021)Arakelyan, Daza, Minervini, and Cochez}]{CQD}
Erik Arakelyan, Daniel Daza, Pasquale Minervini, and Michael Cochez. 2021.
\newblock Complex query answering with neural link predictors.
\newblock In \emph{9th International Conference on Learning Representations,
  {ICLR} 2021, Virtual Event, Austria, May 3-7, 2021}. OpenReview.net.

\bibitem[{Bai et~al.(2022)Bai, Wang, Zhang, and Song}]{Q2P}
Jiaxin Bai, Zihao Wang, Hongming Zhang, and Yangqiu Song. 2022.
\newblock \href {https://doi.org/10.48550/arXiv.2204.12847} {Query2particles:
  Knowledge graph reasoning with particle embeddings}.
\newblock \emph{CoRR}, abs/2204.12847.

\bibitem[{Chen et~al.(2021{\natexlab{a}})Chen, He, Wu, and Wang}]{TACT}
Jiajun Chen, Huarui He, Feng Wu, and Jie Wang. 2021{\natexlab{a}}.
\newblock Topology-aware correlations between relations for inductive link
  prediction in knowledge graphs.
\newblock In \emph{Thirty-Fifth {AAAI} Conference on Artificial Intelligence,
  {AAAI} 2021, Thirty-Third Conference on Innovative Applications of Artificial
  Intelligence, {IAAI} 2021, The Eleventh Symposium on Educational Advances in
  Artificial Intelligence, {EAAI} 2021, Virtual Event, February 2-9, 2021},
  pages 6271--6278. {AAAI} Press.

\bibitem[{Chen et~al.(2022)Chen, Zhang, Zhu, Zhou, Yuan, Xu, and Chen}]{MorsE}
Mingyang Chen, Wen Zhang, Yushan Zhu, Hongting Zhou, Zonggang Yuan, Changliang
  Xu, and Huajun Chen. 2022.
\newblock \href {https://doi.org/10.1145/3477495.3531757} {Meta-knowledge
  transfer for inductive knowledge graph embedding}.
\newblock In \emph{{SIGIR} '22: The 45th International {ACM} {SIGIR} Conference
  on Research and Development in Information Retrieval, Madrid, Spain, July 11
  - 15, 2022}, pages 927--937. {ACM}.

\bibitem[{Chen et~al.(2021{\natexlab{b}})Chen, Hu, and Sun}]{FuzzQE}
Xuelu Chen, Ziniu Hu, and Yizhou Sun. 2021{\natexlab{b}}.
\newblock \href {http://arxiv.org/abs/2108.02390} {Fuzzy logic based logical
  query answering on knowledge graph}.
\newblock \emph{CoRR}, abs/2108.02390.

\bibitem[{Cohen et~al.(2020)Cohen, Yang, and Mazaitis}]{tensorlog}
William~W. Cohen, Fan Yang, and Kathryn Mazaitis. 2020.
\newblock \href {https://doi.org/10.1613/jair.1.11944} {Tensorlog: {A}
  probabilistic database implemented using deep-learning infrastructure}.
\newblock \emph{J. Artif. Intell. Res.}, 67:285--325.

\bibitem[{Gal{\'{a}}rraga et~al.(2015)Gal{\'{a}}rraga, Teflioudi, Hose, and
  Suchanek}]{AMIE+}
Luis Gal{\'{a}}rraga, Christina Teflioudi, Katja Hose, and Fabian~M. Suchanek.
  2015.
\newblock \href {https://doi.org/10.1007/s00778-015-0394-1} {Fast rule mining
  in ontological knowledge bases with {AMIE+}}.
\newblock \emph{{VLDB} J.}, 24(6):707--730.

\bibitem[{Gal{\'a}rraga et~al.(2013)Gal{\'a}rraga, Teflioudi, Hose, and
  Suchanek}]{AMIE}
Luis~Antonio Gal{\'a}rraga, Christina Teflioudi, Katja Hose, and Fabian
  Suchanek. 2013.
\newblock Amie: association rule mining under incomplete evidence in
  ontological knowledge bases.
\newblock In \emph{Proceedings of the 22nd international conference on World
  Wide Web}, pages 413--422.

\bibitem[{Galkin et~al.(2022)Galkin, Zhu, Ren, and Tang}]{GNNQE2}
Mikhail Galkin, Zhaocheng Zhu, Hongyu Ren, and Jian Tang. 2022.
\newblock \href {https://doi.org/10.48550/arXiv.2210.08008} {Inductive logical
  query answering in knowledge graphs}.
\newblock \emph{CoRR}, abs/2210.08008.

\bibitem[{Hamaguchi et~al.(2017)Hamaguchi, Oiwa, Shimbo, and Matsumoto}]{OOKB}
Takuo Hamaguchi, Hidekazu Oiwa, Masashi Shimbo, and Yuji Matsumoto. 2017.
\newblock \href {https://doi.org/10.24963/ijcai.2017/250} {Knowledge transfer
  for out-of-knowledge-base entities : {A} graph neural network approach}.
\newblock In \emph{Proceedings of the Twenty-Sixth International Joint
  Conference on Artificial Intelligence, {IJCAI} 2017, Melbourne, Australia,
  August 19-25, 2017}, pages 1802--1808. ijcai.org.

\bibitem[{Hamilton et~al.(2018)Hamilton, Bajaj, Zitnik, Jurafsky, and
  Leskovec}]{GQE}
William~L. Hamilton, Payal Bajaj, Marinka Zitnik, Dan Jurafsky, and Jure
  Leskovec. 2018.
\newblock Embedding logical queries on knowledge graphs.
\newblock In \emph{Advances in Neural Information Processing Systems 31: Annual
  Conference on Neural Information Processing Systems 2018, NeurIPS 2018,
  December 3-8, 2018, Montr{\'{e}}al, Canada}, pages 2030--2041.

\bibitem[{Hu et~al.(2022)Hu, Guti{\'{e}}rrez{-}Basulto, Xiang, Li, Li, and
  Pan}]{TEMP}
Zhiwei Hu, V{\'{\i}}ctor Guti{\'{e}}rrez{-}Basulto, Zhiliang Xiang, Xiaoli Li,
  Ru~Li, and Jeff~Z. Pan. 2022.
\newblock \href {https://doi.org/10.24963/ijcai.2022/427} {Type-aware
  embeddings for multi-hop reasoning over knowledge graphs}.
\newblock In \emph{Proceedings of the Thirty-First International Joint
  Conference on Artificial Intelligence, {IJCAI} 2022, Vienna, Austria, 23-29
  July 2022}, pages 3078--3084. ijcai.org.

\bibitem[{Kotnis et~al.(2021)Kotnis, Lawrence, and Niepert}]{BiQE}
Bhushan Kotnis, Carolin Lawrence, and Mathias Niepert. 2021.
\newblock Answering complex queries in knowledge graphs with bidirectional
  sequence encoders.
\newblock In \emph{Thirty-Fifth {AAAI} Conference on Artificial Intelligence,
  {AAAI} 2021, Thirty-Third Conference on Innovative Applications of Artificial
  Intelligence, {IAAI} 2021, The Eleventh Symposium on Educational Advances in
  Artificial Intelligence, {EAAI} 2021, Virtual Event, February 2-9, 2021},
  pages 4968--4977. {AAAI} Press.

\bibitem[{Liu et~al.(2021)Liu, Grau, Horrocks, and Kostylev}]{INDIGO}
Shuwen Liu, Bernardo~Cuenca Grau, Ian Horrocks, and Egor~V. Kostylev. 2021.
\newblock {INDIGO:} gnn-based inductive knowledge graph completion using
  pair-wise encoding.
\newblock In \emph{Advances in Neural Information Processing Systems 34: Annual
  Conference on Neural Information Processing Systems 2021, NeurIPS 2021,
  December 6-14, 2021, virtual}, pages 2034--2045.

\bibitem[{Mai et~al.(2021)Mai, Zheng, Yang, and Hu}]{CoMPILE}
Sijie Mai, Shuangjia Zheng, Yuedong Yang, and Haifeng Hu. 2021.
\newblock Communicative message passing for inductive relation reasoning.
\newblock In \emph{Thirty-Fifth {AAAI} Conference on Artificial Intelligence,
  {AAAI} 2021, Thirty-Third Conference on Innovative Applications of Artificial
  Intelligence, {IAAI} 2021, The Eleventh Symposium on Educational Advances in
  Artificial Intelligence, {EAAI} 2021, Virtual Event, February 2-9, 2021},
  pages 4294--4302. {AAAI} Press.

\bibitem[{Meilicke et~al.(2019)Meilicke, Chekol, Ruffinelli, and
  Stuckenschmidt}]{AnyBURL}
Christian Meilicke, Melisachew~Wudage Chekol, Daniel Ruffinelli, and Heiner
  Stuckenschmidt. 2019.
\newblock Anytime bottom-up rule learning for knowledge graph completion.
\newblock In \emph{{IJCAI}}, pages 3137--3143. ijcai.org.

\bibitem[{Ortona et~al.(2018)Ortona, Meduri, and Papotti}]{IEEE2018Rudik}
Stefano Ortona, Venkata~Vamsikrishna Meduri, and Paolo Papotti. 2018.
\newblock Robust discovery of positive and negative rules in knowledge bases.
\newblock In \emph{2018 IEEE 34th International Conference on Data Engineering
  (ICDE)}, pages 1168--1179. IEEE.

\bibitem[{Ren et~al.(2020)Ren, Hu, and Leskovec}]{Q2B}
Hongyu Ren, Weihua Hu, and Jure Leskovec. 2020.
\newblock Query2box: Reasoning over knowledge graphs in vector space using box
  embeddings.
\newblock In \emph{8th International Conference on Learning Representations,
  {ICLR} 2020, Addis Ababa, Ethiopia, April 26-30, 2020}. OpenReview.net.

\bibitem[{Ren and Leskovec(2020)}]{BetaE}
Hongyu Ren and Jure Leskovec. 2020.
\newblock Beta embeddings for multi-hop logical reasoning in knowledge graphs.
\newblock In \emph{Advances in Neural Information Processing Systems 33: Annual
  Conference on Neural Information Processing Systems 2020, NeurIPS 2020,
  December 6-12, 2020, virtual}.

\bibitem[{Sadeghian et~al.(2019)Sadeghian, Armandpour, Ding, and
  Wang}]{sadeghian2019drum}
Ali Sadeghian, Mohammadreza Armandpour, Patrick Ding, and Daisy~Zhe Wang. 2019.
\newblock {DRUM:} end-to-end differentiable rule mining on knowledge graphs.
\newblock In \emph{Advances in Neural Information Processing Systems 32: Annual
  Conference on Neural Information Processing Systems 2019, NeurIPS 2019,
  December 8-14, 2019, Vancouver, BC, Canada}, pages 15321--15331.

\bibitem[{Teru et~al.(2020)Teru, Denis, and Hamilton}]{GraIL}
Komal~K. Teru, Etienne~G. Denis, and William~L. Hamilton. 2020.
\newblock Inductive relation prediction by subgraph reasoning.
\newblock In \emph{Proceedings of the 37th International Conference on Machine
  Learning, {ICML} 2020, 13-18 July 2020, Virtual Event}, volume 119 of
  \emph{Proceedings of Machine Learning Research}, pages 9448--9457. {PMLR}.

\bibitem[{Toutanova et~al.(2015)Toutanova, Chen, Pantel, Poon, Choudhury, and
  Gamon}]{FB15k-237}
Kristina Toutanova, Danqi Chen, Patrick Pantel, Hoifung Poon, Pallavi
  Choudhury, and Michael Gamon. 2015.
\newblock \href {https://doi.org/10.18653/v1/d15-1174} {Representing text for
  joint embedding of text and knowledge bases}.
\newblock In \emph{Proceedings of the 2015 Conference on Empirical Methods in
  Natural Language Processing, {EMNLP} 2015, Lisbon, Portugal, September 17-21,
  2015}, pages 1499--1509. The Association for Computational Linguistics.

\bibitem[{Vaswani et~al.(2017)Vaswani, Shazeer, Parmar, Uszkoreit, Jones,
  Gomez, Kaiser, and Polosukhin}]{transformer}
Ashish Vaswani, Noam Shazeer, Niki Parmar, Jakob Uszkoreit, Llion Jones,
  Aidan~N Gomez, {\L}ukasz Kaiser, and Illia Polosukhin. 2017.
\newblock Attention is all you need.
\newblock In \emph{Advances in neural information processing systems}, pages
  5998--6008.

\bibitem[{Wang et~al.(2019{\natexlab{a}})Wang, Han, Li, and Pan}]{LAN}
Peifeng Wang, Jialong Han, Chenliang Li, and Rong Pan. 2019{\natexlab{a}}.
\newblock \href {https://doi.org/10.1609/aaai.v33i01.33017152} {Logic attention
  based neighborhood aggregation for inductive knowledge graph embedding}.
\newblock In \emph{The Thirty-Third {AAAI} Conference on Artificial
  Intelligence, {AAAI} 2019, The Thirty-First Innovative Applications of
  Artificial Intelligence Conference, {IAAI} 2019, The Ninth {AAAI} Symposium
  on Educational Advances in Artificial Intelligence, {EAAI} 2019, Honolulu,
  Hawaii, USA, January 27 - February 1, 2019}, pages 7152--7159. {AAAI} Press.

\bibitem[{Wang et~al.(2019{\natexlab{b}})Wang, Stepanova, Domokos, and
  Kolter}]{2020_Num}
Po-Wei Wang, Daria Stepanova, Csaba Domokos, and J~Zico Kolter.
  2019{\natexlab{b}}.
\newblock Differentiable learning of numerical rules in knowledge graphs.
\newblock In \emph{International Conference on Learning Representations}.

\bibitem[{Xiong et~al.(2017)Xiong, Hoang, and Wang}]{NELL995}
Wenhan Xiong, Thien Hoang, and William~Yang Wang. 2017.
\newblock \href {https://doi.org/10.18653/v1/d17-1060} {Deeppath: {A}
  reinforcement learning method for knowledge graph reasoning}.
\newblock In \emph{Proceedings of the 2017 Conference on Empirical Methods in
  Natural Language Processing, {EMNLP} 2017, Copenhagen, Denmark, September
  9-11, 2017}, pages 564--573. Association for Computational Linguistics.

\bibitem[{Xu et~al.(2022{\natexlab{a}})Xu, Ye, Chen, Zhao, Chen, and
  Zhang}]{ruleformer}
Zezhong Xu, Peng Ye, Hui Chen, Meng Zhao, Huajun Chen, and Wen Zhang.
  2022{\natexlab{a}}.
\newblock \href {https://doi.org/10.48550/arXiv.2209.05815} {Ruleformer:
  Context-aware differentiable rule mining over knowledge graph}.
\newblock \emph{CoRR}, abs/2209.05815.

\bibitem[{Xu et~al.(2022{\natexlab{b}})Xu, Zhang, Ye, Chen, and Chen}]{ENeSy}
Zezhong Xu, Wen Zhang, Peng Ye, Hui Chen, and Huajun Chen. 2022{\natexlab{b}}.
\newblock \href {https://doi.org/10.48550/arXiv.2209.08779} {Neural-symbolic
  entangled framework for complex query answering}.
\newblock \emph{CoRR}, abs/2209.08779.

\bibitem[{Yang et~al.(2017)Yang, Yang, and Cohen}]{yang2017differentiable}
Fan Yang, Zhilin Yang, and William~W. Cohen. 2017.
\newblock Differentiable learning of logical rules for knowledge base
  reasoning.
\newblock In \emph{Advances in Neural Information Processing Systems 30: Annual
  Conference on Neural Information Processing Systems 2017, December 4-9, 2017,
  Long Beach, CA, {USA}}, pages 2319--2328.

\bibitem[{Yang and Song(2019)}]{2019_NLIL}
Yuan Yang and Le~Song. 2019.
\newblock Learn to explain efficiently via neural logic inductive learning.
\newblock \emph{arXiv preprint arXiv:1910.02481}.

\bibitem[{Zhang et~al.(2021)Zhang, Wang, Chen, Ji, and Wu}]{ConE}
Zhanqiu Zhang, Jie Wang, Jiajun Chen, Shuiwang Ji, and Feng Wu. 2021.
\newblock Cone: Cone embeddings for multi-hop reasoning over knowledge graphs.
\newblock In \emph{Advances in Neural Information Processing Systems 34: Annual
  Conference on Neural Information Processing Systems 2021, NeurIPS 2021,
  December 6-14, 2021, virtual}, pages 19172--19183.

\bibitem[{Zhu et~al.(2022)Zhu, Galkin, Zhang, and Tang}]{GNNQE1}
Zhaocheng Zhu, Mikhail Galkin, Zuobai Zhang, and Jian Tang. 2022.
\newblock Neural-symbolic models for logical queries on knowledge graphs.
\newblock In \emph{International Conference on Machine Learning, {ICML} 2022,
  17-23 July 2022, Baltimore, Maryland, {USA}}, volume 162 of \emph{Proceedings
  of Machine Learning Research}, pages 27454--27478. {PMLR}.

\bibitem[{Zhu et~al.(2021)Zhu, Zhang, Xhonneux, and Tang}]{NBFNet}
Zhaocheng Zhu, Zuobai Zhang, Louis{-}Pascal A.~C. Xhonneux, and Jian Tang.
  2021.
\newblock Neural bellman-ford networks: {A} general graph neural network
  framework for link prediction.
\newblock In \emph{Advances in Neural Information Processing Systems 34: Annual
  Conference on Neural Information Processing Systems 2021, NeurIPS 2021,
  December 6-14, 2021, virtual}, pages 29476--29490.

\end{thebibliography}

\bibliographystylelanguageresource{lrec-coling2024-natbib}
\bibliographylanguageresource{languageresource}

\end{document}